\pdfoutput=1

\documentclass[11pt]{article}

\usepackage{algorithm}
\usepackage{algpseudocode}
\usepackage{amsmath,amssymb}  
\usepackage{bm}               
\usepackage{listings}
\usepackage{tcolorbox}
\usepackage{verbatim}

\usepackage{amsmath}
\usepackage{amssymb}
\usepackage{algorithm}

\usepackage{booktabs}
\usepackage{makecell}
\usepackage{enumitem}
\usepackage{graphicx}
\usepackage{amsmath}
\usepackage{amssymb}

\usepackage{ACL2023}

\usepackage{times}
\usepackage{latexsym}

\usepackage[T1]{fontenc}

\usepackage[utf8]{inputenc}

\usepackage{microtype}

\usepackage{inconsolata}

\title{UIPE: Enhancing LLM Unlearning by Removing Knowledge Related to Forgetting Targets}

\author{
    Wenyu Wang$^1 \footnotemark[1]$, Mengqi Zhang$^1 \footnotemark[1]$, Xiaotian Ye$^2$, Zhaochun Ren$^3$\\
    \textbf{Zhumin Chen}$^{1}\footnotemark[2]$ \ and \ \textbf{Pengjie Ren}$^{1} \footnotemark[2]$ \\ 
    $^1$Shandong University, Qingdao, China \\
    $^2$Beijing University of Posts and Telecommunications, Beijing, China \\
    $^3$Leiden University, Leiden, The Netherlands \\
    {\small \texttt{\{mengqi.zhang, chenzhumin, renpengjie\}@sdu.edu.cn}} \\
    {\small \texttt{z.ren@liacs.leidenuniv.nl}, \texttt{wangwenyu@mail.sdu.edu.cn}, \texttt{yexiaotian@bupt.edu.cn}}
}

\begin{document}
\maketitle
\renewcommand{\thefootnote}{\fnsymbol{footnote}}
\footnotetext[1]{Equal contribution.}
\footnotetext[2]{Corresponding author.}
\begin{abstract}
Large Language Models (LLMs) inevitably acquire harmful information during training on massive datasets. LLM unlearning aims to eliminate the influence of such harmful information while maintaining the model's overall performance. Existing unlearning methods, represented by gradient ascent-based approaches, primarily focus on forgetting target data while overlooking the crucial impact of logically related knowledge on the effectiveness of unlearning. In this paper, through both theoretical and experimental analyses, we first demonstrate that a key reason for the suboptimal unlearning performance is that models can reconstruct the target content through reasoning with logically related knowledge. To address this issue, we propose Unlearning Improvement via Parameter Extrapolation (UIPE), a method that removes knowledge highly correlated with the forgetting targets. Experimental results show that UIPE significantly enhances the performance of various mainstream LLM unlearning methods on the TOFU benchmark.
\end{abstract}

\section{Introduction}
Large language models (LLMs) trained on massive datasets show exceptional capabilities \citep{kaplan2020scaling,wei2022emergent}. However, such extensive datasets inevitably contain harmful information, which diminishes model performance and may cause societal challenges. \citep{yao2024machine}. For instance, LLMs expose private information, copyrighted content and inherent biases from their training data \citep{carlini2021extracting,huang2022large,zhao2024gender}.

To address the aforementioned risks, LLM unlearning has emerged as a critical research direction. LLM unlearning aims to mitigate the influence of undesired data \citep{cao2015towards,liu2024rethinking,wang2023kga,eldan2023s,liu2024towards}. Gradient ascent-based (GA) LLM unlearning has emerged as one of the predominant methodologies in this field \citep{jang2022knowledge}.

\begin{figure}[t!]
\centering
\includegraphics[width=\linewidth]{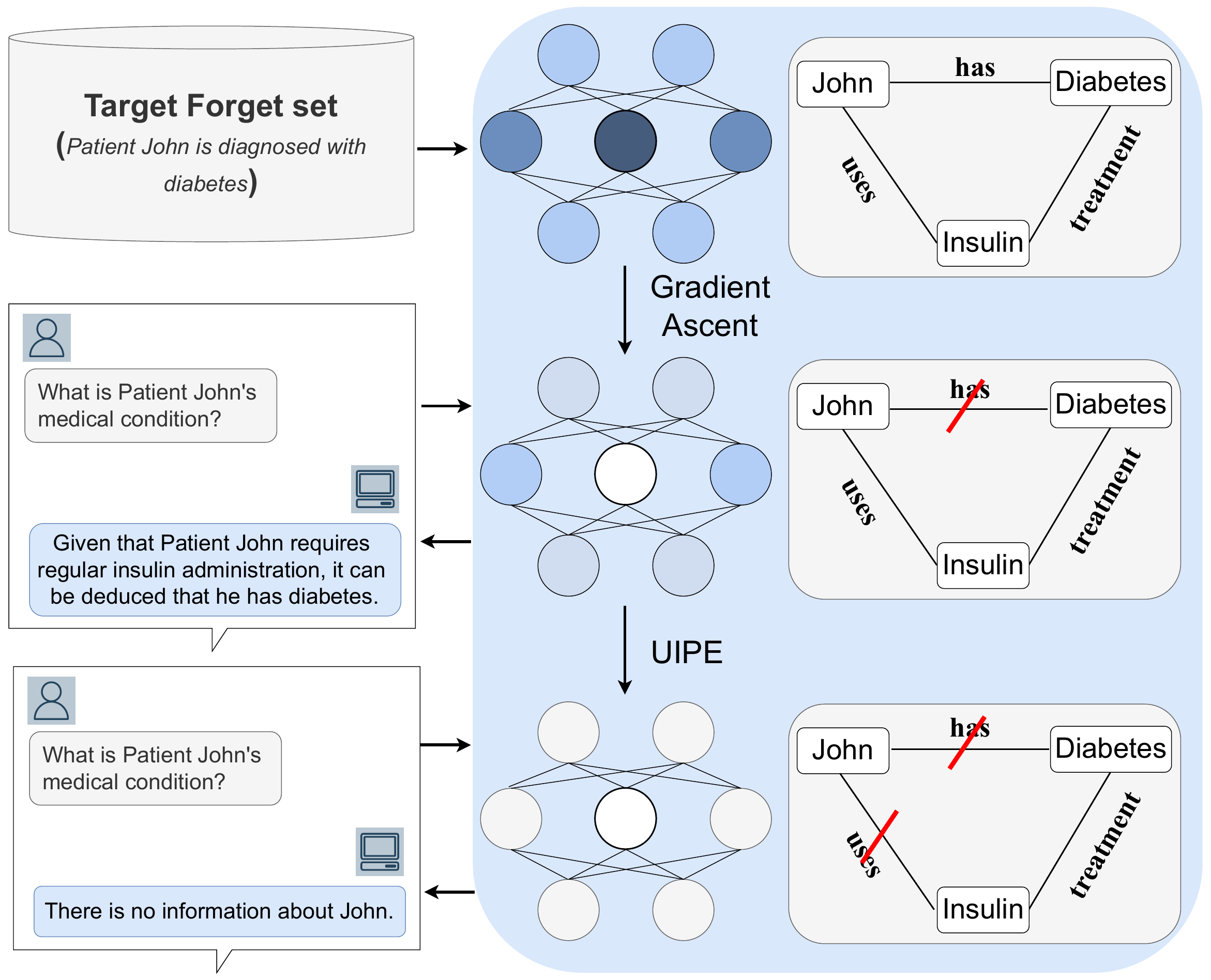}
\vspace{-5pt}
\caption{UIPE is motivated by the observation that  after gradient ascent unlearning of John's private data, the model still retains logically related knowledge, allowing it to infer the forgotten information.}
\label{intro}
\end{figure}

Recently, numerous studies have emerged aimed at improving the GA method. A popular approach regularizes the objective by combining forgetting and utility losses, aiming to forget specific data while preserving performance, such as Grad. Diff. \citep{yao2023large} and KL Min. \citep{chen2023unlearn}. Additionally, inspired by Direct Preference Optimization (DPO) \citep{rafailov2024direct}, negative preference optimization (NPO) alleviates catastrophic collapse during the forgetting process \citep{zhang2024negative}. Despite these advancements, effective unlearning techniques for LLMs remain an open challenge \citep{maini2024tofu,choi2024breaking,shumailov2024ununlearning}.

We propose the hypothesis that one of the key factors contributing to the suboptimal unlearning performance of LLMs is that they can infer the knowledge that should have been forgotten by leveraging logically related knowledge. For instance, as shown in Figure \ref{intro}, even if a model forgets the knowledge \textit{``Patient John is diagnosed with diabetes''} from the target forget set, it may still reconstruct this knowledge through related knowledge outside the target forget dataset, such as \textit{``Patient John requires regular insulin administration''} and \textit{``Insulin is a standard treatment for diabetes management''}.

To validate our hypothesis, we conduct a preliminary experiment using a virtual character dataset, which contains both a target forget set and a related knowledge set (\S \ref{pre-experiment}). Our results reveal that when a model is trained on both sets, unlearning only the target forgetting set is insufficient for complete knowledge removal. However, when related knowledge is included in the unlearning process, the model demonstrates significantly improved forgetting effectiveness on the target forget set. These findings suggest that LLMs can reconstruct target knowledge that should be forgotten by related information.

Given that LLMs are trained on massive datasets, and their training data is often inaccessible, constructing complete related knowledge sets remains a major challenge. This raises a crucial question: \emph{Can related knowledge unlearning be achieved without requiring additional training data?} To address this, we propose UIPE (Unlearning Improvement via Parameter Extrapolation), a plug-and-play auxiliary unlearning method (\S \ref{method}). This method is founded on a crucial observation: the unlearning of target knowledge triggers the forgetting of related knowledge. This phenomenon stems from the fact that related knowledge exhibits similar distribution characteristics in the parameter space, leading to highly correlated gradient changes \citep{qin2024does,xie2024gradsafe}. By amplifying the gradient ascent updates on the target forget set, we extend its gradient update effects to the related knowledge set, significantly enhancing the model's capability to forget related knowledge. Experimental evaluations based on the TOFU benchmark demonstrate that our method enables various unlearning approaches to achieve optimal trade-offs between forget quality and model utility preservation.

We summarize our {contributions} below.
\begin{itemize}
    \item We identify the limitation of the GA method in unlearning related knowledge, which we found to be a key factor behind the unsatisfactory unlearning performance of models.
\end{itemize}
\begin{itemize}
    \item We introduce the UIPE method, which utilizes parameter extrapolation to enhance the model's ability to forget related knowledge.
\end{itemize}
\begin{itemize}
    \item  We conduct experiments on various GA-based unlearning methods using the TOFU benchmark. The results demonstrate that UIPE facilitates a more optimal balance between model utility and forget quality across these methods.
\end{itemize}

\section{Related Work}
\subsection{Machine unlearning}
Machine unlearning, a concept rooted in data protection regulations like the `right to be forgotten' \cite{rosen2011right}, has evolved beyond its initial scope of general data protection frameworks \cite{cao2015towards,hoofnagle2019european,bourtoule2021machine,nguyen2022survey}. 
The field has experienced rapid expansion, with applications now spanning multiple domains, including image classification \cite{ginart2019making,golatkar2020eternal,kurmanji2024towards,jia2023model}, generative AI tasks such as text-to-image and image-to-image synthesis \cite{zhang2023forget,kumari2023ablating,gandikota2023erasing,fan2023salun,li2024machine}, and federated learning systems \cite{wang2022federated, liu2023survey}.

In the research literature, `exact' unlearning refers to the complete retraining of a model while excluding the designated forgotten data points \cite{nguyen2022survey,jia2023model,fan2024challenging}. However, this approach has practical limitations due to high computational costs and data access requirements, leading to the development of more efficient `approximate' unlearning methods \cite{golatkar2020eternal,graves2021amnesiac,chen2023boundary,kurmanji2024towards,jia2023model}. Furthermore, several methodologies now offer provable and certified data removal guarantees \cite{guo2019certified,ullah2021machine,sekhari2021remember}.

\subsection{LLM unlearning}
The importance of unlearning in LLLMs has increasingly emerged, attracting more and more attention \citet{liu2024rethinking,zhang2023right}. Several research efforts have focused on employing gradient ascent techniques to achieve forgetting in target datasets \cite{jang2022knowledge, yao2023large,chen2023unlearn,maini2024tofu,zhang2024negative}. Meanwhile, WHP and its improved variant construct the teacher distribution through a name replacement strategy to achieve the goal of forgetting target knowledge \cite{eldan2023s,liu2024revisiting}. SOUL investigated the impact of second-order optimizers on unlearning effectiveness \citet{jia2024soul}. Some unlearning methods have explored the data-model interactions that could influence LLM unlearning, such as weight localization-based unlearning \cite{yu2023unlearning,jia2024wagle}, achieving forgetting through modifications to LLMs' hidden representations \cite{li2024wmdp} or perturbations to the model's embedding layer \cite{liu2024large}. Additionally, ULD achieved unlearning through an auxiliary smaller model \citet{ji2024reversing}. Finally, researchers have developed several benchmarks for evaluating LLM unlearning effectiveness, such as TOFU for fictitious unlearning \cite{maini2024tofu}, WMDP for unlearning hazardous knowledge in LLMs \cite{li2024wmdp} and RWKU  for zero-shot konwledge unlearning \cite{jin2024rwku}.

\section{Preliminaries}

\subsection{Unlearning}
\textbf{LLM unlearning} strives to eliminate undesired data without significantly compromising the overall performance of large language models. We represent question-answer pairs derived from specific factual knowledge $k_i$ as $(x_i,y_i)$, where $x_i$ denotes the question and $y_i$ represents the corresponding answer. Given a dataset $\mathcal{D} = \left\{ (x_i,y_i) \right\}_{i = 1}^{n}$ containing $n$ question-answer pairs, let $\mathcal{P}_{\theta}$ be a model trained on $\mathcal{D}$. The goal of LLM unlearning is to ensure that $\mathcal{P}_{\theta}$ completely forgets the knowledge contained in the target forget set $\mathcal{D}_{f} = \left\{ (x_i,y_i) \right\}_{i = 1}^{m}$ ($m<n$). After unlearning, the model's performance should be indistinguishable from a model trained exclusively on the retained dataset $\mathcal{D}_{r}=\mathcal{D} \backslash \mathcal{D}_{f}$.

\noindent \textbf{Evaluation} of LLM unlearning effectiveness is typically assessed along two key dimensions \cite{maini2024tofu}: model utility, which measure the general capabilities of the unlearned model, and forget quality, which quantifies the extent to which the targeted knowledge has been successfully removed.

\noindent \textbf{Gradient ascent} is an important method for LLM unlearning, designed to reverse the optimization process on a designated forget set. The method builds upon the standard training paradigm of the $\mathcal{P}_{\theta}$, which minimizes the prediction loss over the full dataset $\mathcal{D}$. To enforce forgetting, gradient ascent maximizes the prediction loss on the target forget subset $\mathcal{D}_{f}$, effectively approximating the reversal of the original optimization process. This procedure can be equivalently interpreted as performing gradient descent on the negative prediction loss \cite{zhang2024negative}. The gradient ascent objective, denoted as $\mathcal{L}_{GA}$, is formulated as:
\begin{align}
\mathcal{L}_{GA}(\theta) = \mathbb{E}_{\mathcal{D}_{f}}\left\lbrack {\log\left( {\mathcal{P}_{\theta}\left( y \middle| x \right)} \right)} \right\rbrack .
\end{align}

\subsection{Similar Parameter Distribution of Related Knowledge} \label{3.2}
In this paper, related knowledge refers to knowledge that is logically connected to a target piece of knowledge and can be used to infer or reconstruct it. Even after direct unlearning, an LLM may still recall forgotten information by leveraging related knowledge. Formally, given a knowledge instance $k_i$ in the target forget set, another knowledge instance $k_i^{\prime}$ is considered related knowledge if the model can logically derive $k_i$ from $k_i^{\prime}$ using its internal reasoning mechanisms.

In LLMs, related knowledge typically exhibits similar storage distribution patterns, leading to correlated parameter updates during model training \cite{qin2024does}.When modeling the storage characteristics of $k_i$ and $k_i'$ in the model through gradients, these related knowledge instances often demonstrate high cosine similarity in their gradients. For example, consider two related question-answer pairs: based on knowledge $k_i$, the pair $(x_i,y_i)$ consists of "\textit{What is patient John's condition?}" and "\textit{Patient John has been diagnosed with diabetes.}", while based on knowledge $k_i'$, the pair $(x_i',y_i')$ consists of "\textit{What treatment did John receive?}" and "\textit{Patient John requires regular insulin injections.}". When modeling the storage distribution of $k_i$ and $k_i'$ using gradients, their respective gradients $\nabla_{\theta}\mathcal{P}_{\theta}\left( y_{i}| x_{i} \right)$ and $\nabla_{\theta}\mathcal{P}_{\theta}\left( y_i'| x_i' \right)$ exhibit high cosine similarity, indicating their interdependence. This similarity is quantified as:
\begin{align}
\mathcal{R}_{\theta}( k_{i},k_{i}^{'}) = \cos\left( \nabla_{\theta}\mathcal{P}_{\theta}\left( y_{i}| x_{i} \right),\nabla_{\theta}\mathcal{P}_{\theta}\left( y_i'| x_i' \right) \right)  \label{cosine similarity}
\end{align}

\section{Preliminary Experiments}
\label{pre-experiment}
To validate this hypothesis that LLMs can leverage related knowledge to reconstruct forgotten knowledge, we first construct a target forget set along with a corresponding related knowledge set, and then conduct a series of comparative experiments to systematically evaluate this phenomenon.
\subsection{Data Construction and Evaluation Metrics}\label{Data Construction and Evaluation Metrics}
We construct a comprehensive synthetic personal dataset comprising two subsets: a target forget set and a related knowledge set. Specifically, we utilize GPT-4 to generate experimental data for 12 fictional individuals, each characterized by $10$ specific attributes (e.g., biometric features, address, etc.). For each attribute, we meticulously design two corresponding question-answer pairs: $(x_i,y_i)$ explicitly describes the personal information associated with the attribute, while $(x_i', y_i')$ is logically related to $(x_i, y_i)$, and can be inferred from it based on the model's inherent common-sense reasoning capabilities. 

As a result, the collection of all $(x_i,y_i)$ pairs constitutes the target forget set, while all corresponding $(x_i',y_i')$ pairs form the related knowledge set. Notably, all data in this dataset are entirely synthetic, ensuring that the model has not been exposed to this information during pre-training. Detailed prompts and data samples are provided in Appendix \ref{Prompt and Data Sample}.

To assess the effectiveness of unlearning, we evaluate model utility using ROUGE-L \citep{lin-2004-rouge} scores on the TruthfulQA \citep{lin-etal-2022-truthfulqa} dataset. Meanwhile, we measure forget quality by computing ROUGE-L scores on the target forget set.

\subsection{Impact of Related Knowledge on LLM Unlearning}
In this experiment, we investigate the influence of related knoweldge on the effectiveness of unlearning in LLMs, using LLaMA-2-7b-chat \cite{touvron2023llama} as the research subject. By applying different combinations of training data and unlearning operations, we construct multiple model variants to systematically analyze how related knowledge affects the unlearning process. Table \ref{tab:preliminary} provides the detailed experimental configurations.  

\begin{itemize}[leftmargin=*]
    \item We first fine-tune the LLaMA-2-7b-chat on both the target forget set and related knowledge set, allowing it internalize all relevant knowledge. We then apply the GA method to unlearn only the target forget set, resulting in model $\mathcal {P}_{\theta_{1}}$. It simulates the unlearning process in real scenarios.
    
    \item We fine-tune the LLaMA-2-7b-chat exclusively on the target forget set, ensuring it has no prior exposure to related knowledge.  We then apply the GA method to unlearn the target forget set, yielding model $\mathcal {P}_{\theta_{2}}$. 
    
    \item We fine-tune the model on both the target forget set and related knowledge set. We then employ the GA method to simultaneously unlearn both knowledge sets, producing model $\mathcal {P}_{\theta_{3}}$. This setup allows us to investigate whether explicitly unlearning related knowledge improves the effectiveness of forgetting the target knowledge.
\end{itemize}

\begin{table}[t!]
    \centering
    \caption{Variant Models with their corresponding training data and unlearning operations.}
\resizebox{\linewidth}{!}
{\begin{tabular}{lcc}
        \toprule
        Model & Fine-Tune Dataset & Unlearning Dataset \\
        \midrule
        $\mathcal {P}_{\theta_{1}}$ & \makecell{target forget set \\ related knowledge set} & \makecell{target forget set} \\
        \midrule
        $\mathcal {P}_{\theta_{2}}$ & \makecell{target forget set} & \makecell{target forget set} \\
        \midrule
        $\mathcal {P}_{\theta_{3}}$ & \makecell{target forget set \\ related knowledge set} & \makecell{target forget set \\ related knowledge set} \\
        \bottomrule
    \end{tabular}}  
    \label{tab:preliminary}
\end{table}

\begin{figure}[ht]
\centering
\includegraphics[width=\linewidth]{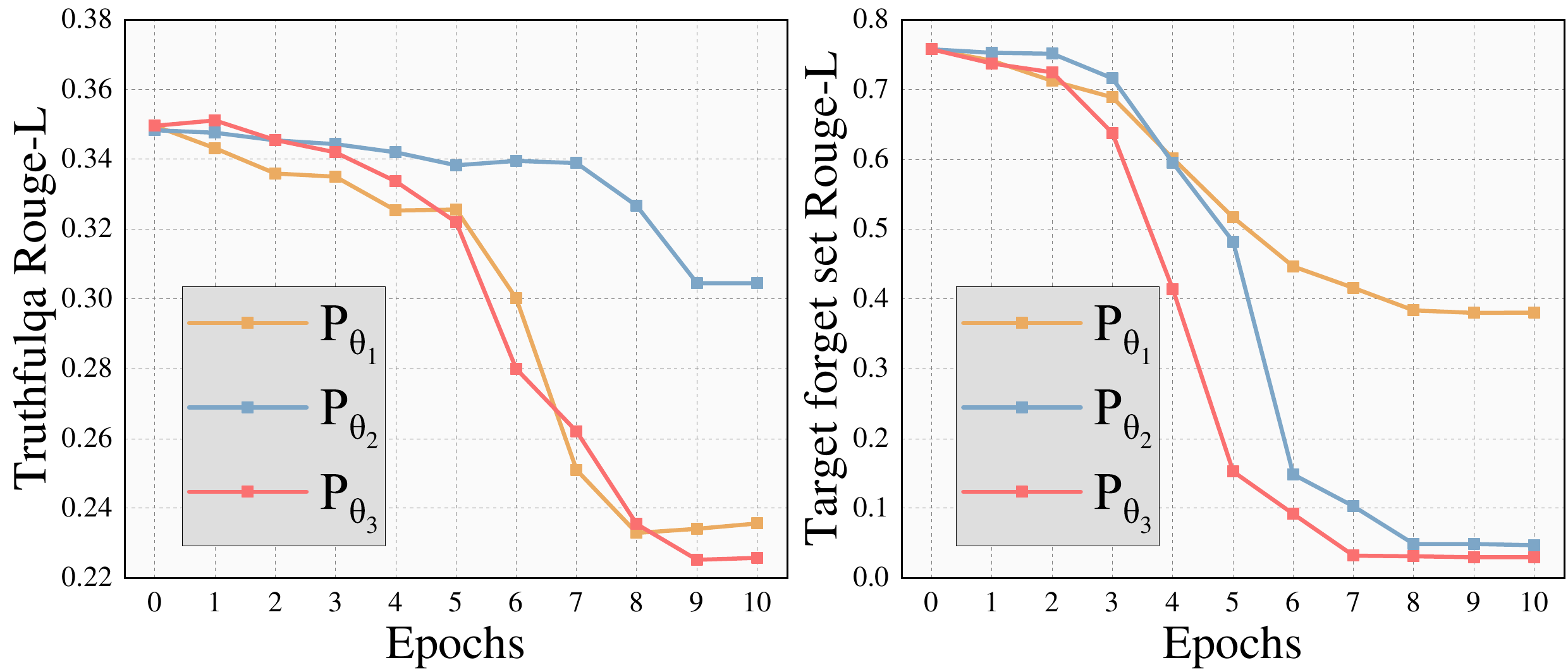}
\vspace{-5pt}
\caption{Model unlearning performance over 10 epochs. Left: Model utility (higher Rouge-L score indicates better utility). Right: Forget quality (lower Rouge-L score indicates unlearning effectiveness).}
\label{fig:experiment1,2,3}
\end{figure}
Figure \ref{fig:experiment1,2,3} presents the performance of the models during the unlearning process across different epochs, evaluating both forget quality and model utility. From the results, we can draw the following conclusions:
\begin{itemize}[leftmargin=*]
    \item \textbf{Models can reconstruct forgotten knowledge by leveraging related knowledge}. Compared to $\mathcal {P}_{\theta_{2}}$, $\mathcal {P}_{\theta_{1}}$ exhibits poorer model utility and lower forget quality. The key difference between these models is $\mathcal {P}_{\theta_{1}}$ was trained on both the target forget set and the related knowledge set, whereas $\mathcal {P}_{\theta_{2}}$ was trained only ont the target forget set. Consequently, even after unlearning the target forget set, $\mathcal {P}_{\theta_{1}}$ can still reconstruct the forgotten knowledge by leveraging related knowledge, leading to suboptimal forgetting performance.  This finding validates our hypothesis that related knowledge enables LLMs to infer forgotten information, reducing the effectiveness of unlearning. 
    
    \item \textbf{Unlearning related knowledge enhances forget quality on the target forget set}. Compared to $\mathcal {P}_{\theta_{1}}$, $\mathcal {P}_{\theta_{3}}$, which undergoes unlearning on both the target forget set and the related knowledge set, demonstrates a significant improvement in forget quality on the target forget set, while maintaining comparable model utility. This further validates the correctness of our hypothesis.
\end{itemize}

Despite these findings, real-world application remains challenging. The vast scale of LLM training data and the difficulty of identifying internal knowledge make constructing a comprehensive related knowledge set infeasible. As a result, replicating the approach used for $\mathcal {P}_{\theta_{3}}$, where both target and related knowledge are unlearned—is impractical. This raises a critical question: \textbf{Can related knowledge be unlearned without additional training data?}


\section{Methodology}
\label{method}
\subsection{Rethinking the Effectiveness of GA}
To address the existing challenge, we conduct an in-depth analysis of model $\mathcal{P}_{\theta_{1}}$, which is first fine-tune on both the target forget set and related knowledge set, followed by unlearning on the target forget set. During the inference phase, we evaluate not only the forget quality on the target forget set but also evaluate its forget quality on the related knowledge set, thereby systematically analyzing the forgetting effects of $\mathcal{P}_{\theta_{1}}$ on both datasets.
\begin{figure}[ht]
\centering
\includegraphics[width=0.85\linewidth]{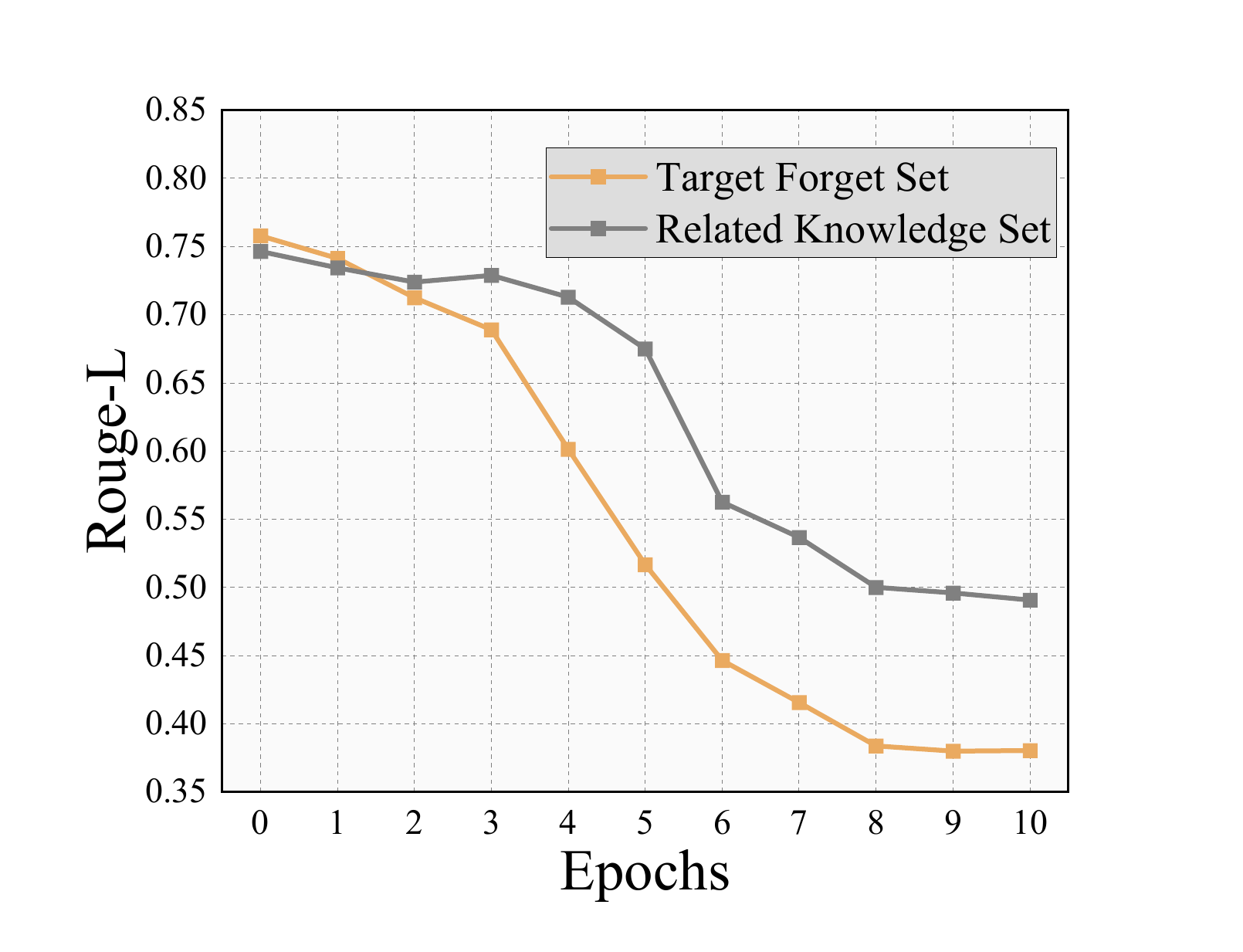}
\vspace{-5pt}
\caption{$\mathcal {P}_{\theta_{1}}$'s forget quality on both the target forget set and the related knowledge set, unlearning for 10 epochs (lower Rouge-L score indicates better quality).}
\label{fig:k1,k2 test}
\end{figure}

Through Figure \ref{fig:k1,k2 test}, we observe an interesting phenomenon: although $\mathcal {P}_ {\theta_{1}}$ only undergoes unlearning training on the target forget set, it improves the forget quality not only for the target forget set but also for the related knowledge set.

\textbf{We first analyze how the GA method facilitates the forgetting of target knowledge.} Formally, we use $\mathcal {P}_ {\theta_{ini}}$ denote the initial model corresponding to $\mathcal {P}_ {\theta_{1}}$ that has only undergone fine-tune without unlearning training. For any example $k_{i}=\left(x_{i}, y_{i}\right)$ in the target forget set (its corresponding example $k_i'=\left(x_{i}', y_{i}'\right)$ in the related knowledge set), the GA method performs gradient ascent on model $\mathcal {P}_ {\theta_{ini}}$, with the parameter update expressed as:
\begin{align}
\theta_{1} &= \theta_{ini} + \eta \cdot \nabla_{\theta}\mathcal{L}_{GA}\left( \theta_{ini} \right) \nonumber \\ 
&= \theta_{ini} + \underbrace {\eta \cdot  \frac{\nabla_{\theta}\mathcal{P}_{\theta_{ini}}\left( y_{i} \middle| x_{i} \right)}{\mathcal{P}_{\theta_{ini}} (y_{i}|x_{i})}}_{v}
\label{sft2un}
\end{align}
where vector $v$ represents the parameter update of model $\mathcal {P}_ {\theta_{ini}}$ on $k_{i}$, $\nabla_\theta\mathcal{P}_{\theta_{ini}}\left( y_{i} \middle| x_{i} \right)$ is the gradient of $k_{i}$ in the model and $\eta$ is the learning rate. Namely, $\theta_{ini}$ is updated in the direction of \(\nabla_\theta\mathcal{P}_{\theta_{ini}}\left( y_{i} \middle| x_{i} \right)\). Therefore, when the model updates its parameters along the gradient direction of the knowledge in the model, it leads to the forgetting of this knowledge.

\begin{figure}[ht]
\centering
\includegraphics[width=0.9\linewidth]{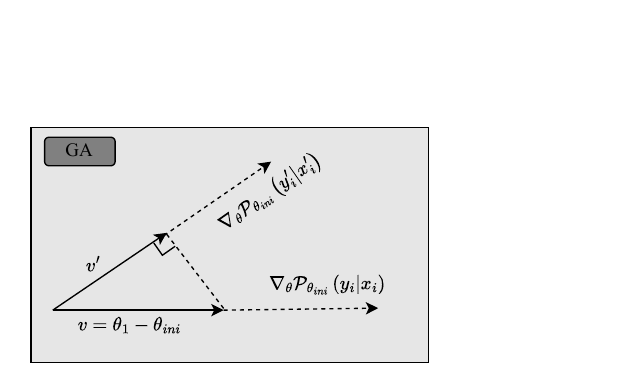}
\vspace{-5pt}
\caption{The parameter update vector \(v\) in the gradient direction of \(k_{i}\) also induces a projected update \({v}'\) in the gradient direction of \(k_{i}'\).}
\label{methodGA}
\end{figure}

\textbf{Furthermore, we analyze how GA is capable of forgetting related knowledge.} Based on the theory of related knowledge sharing similar parameter distributions, we model the storage distributions of knowledge $k_{i}$ and $k_{i}'$ using the gradients $\nabla_\theta\mathcal{P}_{\theta_{ini}}\left( y_{i} \middle| x_{i} \right)$ and $\nabla_\theta\mathcal{P}_{\theta_{ini}}\left( y_{i}' \middle| x_{i}' \right)$ in the model $\mathcal{P}_{\theta_{ini}}$. Since $v$ and $\nabla_\theta\mathcal{P}_{\theta_{ini}}\left( y_{i} \middle| x_{i} \right)$ share the same direction, the cosine similarity $\mathcal{R}_{\theta_{ini}}( k_{i},k_{i}')$ between $\nabla_\theta\mathcal{P}_{\theta_{ini}}\left( y_{i} \middle| x_{i} \right)$ and $\nabla_\theta\mathcal{P}_{\theta_{ini}}\left( y_{i}' \middle| x_{i}' \right)$ is also the cosine similarity between $v$ and $\nabla_\theta\mathcal{P}_{\theta_{ini}}\left( y_{i}' \middle| x_{i}' \right)$. This results in $v$ having a projection component in the direction of $\nabla_\theta\mathcal{P}_{\theta_{ini}}\left( y_{i}' \middle| x_{i}' \right)$, as illustrated in Figure \ref{methodGA}, denoted as ${v}'$. The expression for ${v}'$ can be derived using the projection formula as follows:
\begin{align}
    {v}' = |{v}| \cdot \mathcal{R}_{\theta_{ini}}( k_{i},k_{i}') \cdot {v}'_{o}
    \label{projection}
\end{align}
where ${v}'_{o}$ is the unit vector of $\nabla_\theta\mathcal{P}_{\theta_{ini}}\left( y_{i}' \middle| x_{i}' \right)$. Therefore, the update of the model parameters also generates a projection component in the direction of the gradient of the related knowledge, leading to the forgetting of that knowledge.

However, updates through the projection relationship are limited. As shown in Figure \ref{fig:k1,k2 test}, the forgetting quality on the related knowledge set stops improving towards the end of the unlearning process. Specifically, once the model $P_{\theta_{ini}}$ has completely forgotten knowledge $k_i$, $\nabla_\theta\mathcal{P}_{\theta_{ini}}\left( y_{i} \middle| x_{i} \right)$ no longer represents the storage of $k_i$ in $P_{\theta_{ini}}$. Consequently, $\mathcal{R}_{\theta_{ini}}( k_{i},k_{i}')$ becomes meaningless, causing the projection relationship in Equation \ref{projection} to fail. This prevents parameter updates in the gradient direction of knowledge $k_i'$, thus making it impossible to continue forgetting knowledge $k_i'$.

\subsection{UIPE} \label{UIPE method}
Based on the observation that model unlearning on the target forget triggers unlearning effects in the related knowledge, we leverage the projection relationship between $v$ and $v'$ to achieve related knowledge unlearning without additional data, thereby proposing the UIPE method.

\begin{figure}[t!]
\centering
\includegraphics[width=0.9\linewidth]{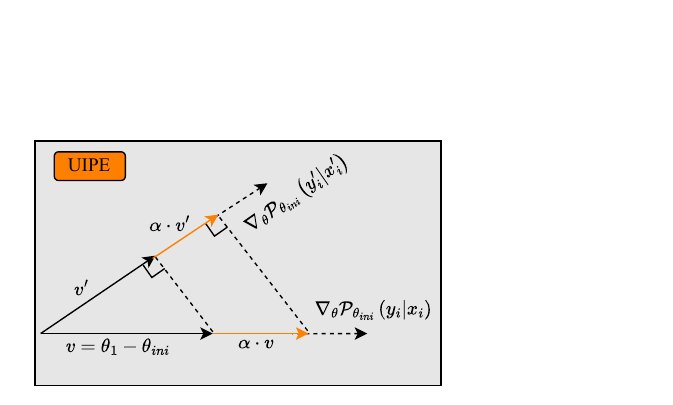}
\vspace{-5pt}
\caption{UIPE amplifies the existing parameter update \(v\) through linear extrapolation, correspondingly amplifying the projection \({v}'\).}
\label{UIPE fig}
\end{figure}

Specifically, we aim to extrapolate the existing parameter update \(v\) made on \(k_i\). Correspondingly, the existing update of the projection \(v'\) in the direction of \(\nabla_\theta\mathcal{P}_{\theta_{ini}}\left( y_{i}' \middle| x_{i}' \right)\) is also extrapolated to achieve more thorough forgetting of the related knowledge. In this paper, we utilize linear extrapolation (as illustrated in Figure \ref{UIPE fig}, simply 
 amplifying the existing updates). The UIPE method can be expressed as:
\begin{align}
    \theta_{uipe} &= \theta_{ini} + ( 1 + \alpha) \cdot v \label{equation:UIPE}
\end{align}
where $\alpha$ is an amplify coefficient controlling the amplification magnitude of ${v}$. This formula shows that compared to the original gradient ascent update \ref{sft2un}, the UIPE method adds an amplified update vector $( 1 + \alpha) \cdot v$ to the initial model parameters $\theta_{ini}$, with the amplification degree controlled by the scalar $\alpha$.
Based on Equation \ref{projection}, the projection of the amplified update vector $( 1 + \alpha) \cdot v$ in the direction of $\nabla_\theta\mathcal{P}_{\theta_{ini}}\left( y_{i}' \middle| x_{i}' \right)$ can be expressed as:
\begin{align}
  (1+\alpha) \cdot {v}' = |(1+\alpha) \cdot v| \cdot \mathcal{R}_{\theta_{ini}}( k_{i},k_{i}') \cdot {v}'_{o}
\end{align}
UIPE increases the model's parameter updates in the direction of $\nabla_\theta\mathcal{P}_{\theta_{ini}}\left( y_{i}' \middle| x_{i}' \right)$ by amplifying $v$. More importantly, due to the presence of $\mathcal{R}_{\theta_{ini}}( k_{i},k_{i}')$, when the update vector $v$ is amplified by a fixed coefficient $\alpha$, UIPE performs larger parameter updates in the corresponding direction for knowledge $k_i'$ that exhibits stronger correlation with knowledge $k_i$ (higher values of $\mathcal{R}_{\theta_{ini}}( k_{i},k_{i}')$).

In practical applications, UIPE can be implemented through three core steps: First, based on the target forget dataset $\mathcal{D}_f$, the initial model $\mathcal{P}_{\theta_{\text{ini}}}$ is trained for multiple rounds using gradient ascent algorithm or its variants. The unlearning model $\mathcal{P}_{\theta_{\text{un}}}$ from the optimal round is selected based on forget quality and model utility, ensuring effective forgetting of target knowledge while maintaining general model capabilities. Next, we compute the parameter update vector $v = \theta_{\text{un}} - \theta_{\text{ini}}$ generated during the unlearning process. Finally, by introducing a hyperparameter $\alpha$ to directionally amplify $v$, we add the extrapolated update $\alpha \cdot v$ to $\theta_{\text{un}}$, enhancing the model's ability to forget knowledge highly related with the target knowledge, ultimately outputting the optimized model $\mathcal{P}_{\theta_{\text{uipe}}}$.

\section{Experiments}
\begin{figure*}[t!]
    \centering
    \includegraphics[width=\linewidth]{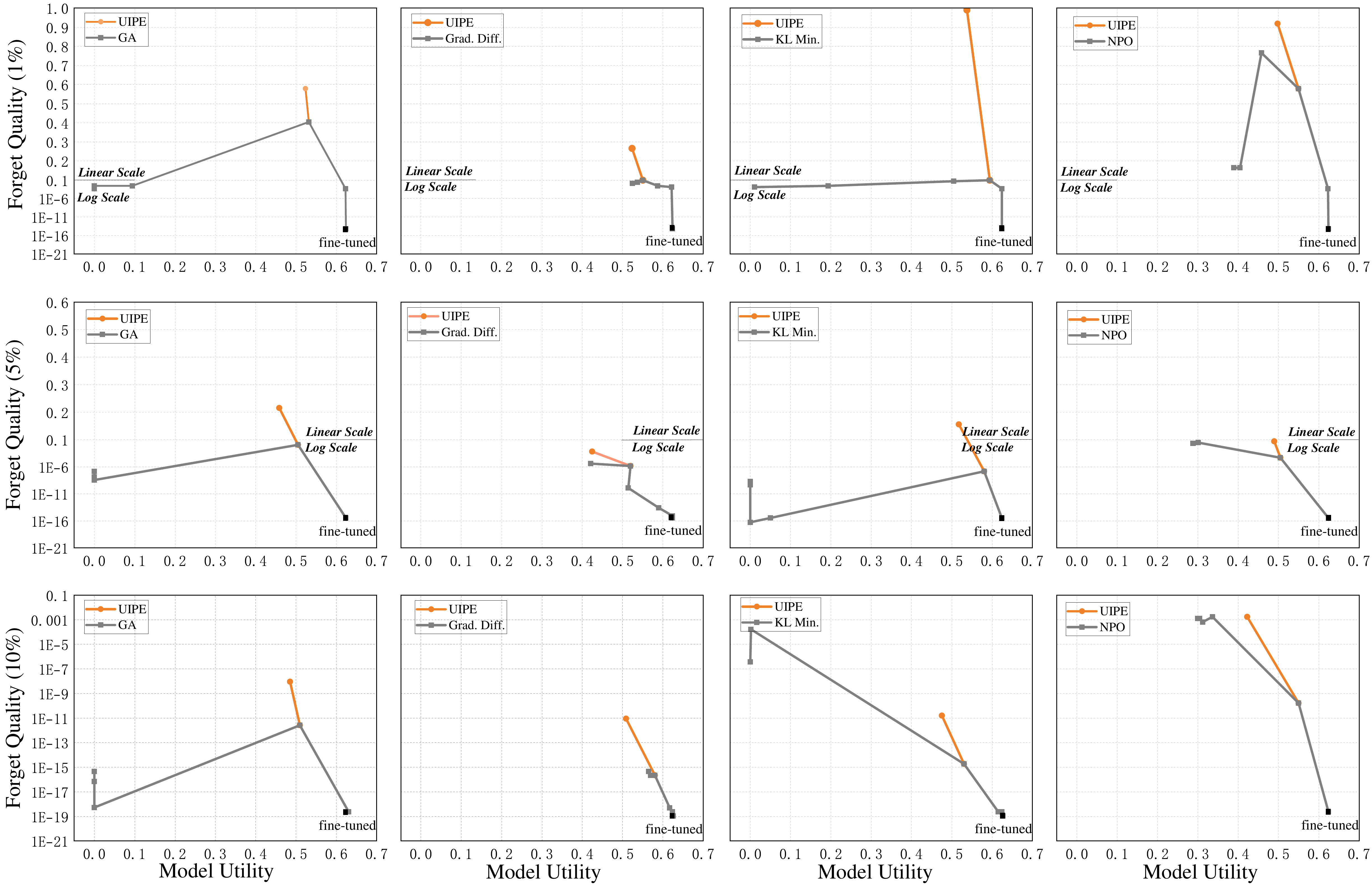}
    \caption{Results of TOFU benchmark tests after applying UIPE to four baseline LLM unlearning methods. For the 1\% and 5\% target forget datasets, dual-scale plots are employed (linear scale above and logarithmic scale below the black line), while the 10\% dataset uses a uniform logarithmic scale throughout. Gray lines illustrate the baseline method trajectories (black dots indicate initial metrics, gray dots show metrics after five unlearning epochs), while orange lines represent metric changes after UIPE application.}
    \label{main result}
    \vspace{-2pt}
\end{figure*}

\subsection{Experimental setup}
\textbf{Dataset and Model}. We assess the performance of UIPE on the TOFU benchmark \cite{maini2024tofu}, which includes 200 fictional author profiles, each containing $20$ question-answer pairs. TOFU defines three forgetting levels: Forget01, Forget05, and Forget10, which correspond to the forgetting of $1\%$, $5\%$, and $10\%$ of the data, respectively. The effectiveness of the unlearning methods is evaluated on the LLaMA-2-7B-chat model using two metrics: Forget Quality and Model Utility, as described in \citet{maini2024tofu}. 

\noindent \textbf{Baselines}. We evaluate the effectiveness of the proposed UIPE method by applying it to a series of LLM unlearning techniques. In addition to the basic GA method, we conduct experiments with Grad. Diff. \citep{yao2023large},KL Min. \citep{chen2023unlearn}, and NPO \citep{zhang2024negative} using the TOFU benchmark. Detailed descriptions of these methods are provided in the Appendix \ref{Baseline LLM unlearning methods}.

Typically, we select the epoch with optimal forget quality from the baseline methods to apply UIPE. However, when the model with optimal forget quality exhibits low model utility, improving its forget quality becomes meaningless. In response, we opt for models with higher utility but sub-optimal forget quality. Experimental results demonstrate that this strategy effectively achieves an optimal trade-off between forget quality and model utility.

\subsection{Results} \label{Results}

\textbf{UIPE helps baseline unlearning methods achieve optimal trade-offs in most scenarios}. Figure \ref{main result} illustrates the improvements made by UIPE on the trade-off between forget quality and model utility for various unlearning methods in Forget01, Forget05, and Forget10. Specifically, GA, Grad.Diff., and KL Min. methods demonstrate substantial improvements in forgetting performance during the initial phase. However, these methods show suboptimal performance in subsequent updates: GA and KL Min. suffer from significant drops in model utility, while Grad.Diff. experiences poor forget quality. This indicates that continuing unlearning training with these methods fails to effectively enhance the model's forgetting performance. In contrast, when combined with UIPE, these methods show marked improvements. Notably, for the Forget01 dataset, UIPE helps KL Min. achieve near-ideal forget quality (1.0) with minimal loss in model utility. Although NPO significantly outperforms the other three baseline methods, UIPE further enhances its forgetting performance. For the Forget01 dataset, UIPE enables NPO to reach a new optimal forget quality while effectively reducing model utility loss. On Forget05 and Forget10 datasets, while UIPE does not surpass NPO's best forget quality, it maintains high forget quality while significantly reducing model utility loss.

As the scale of forgetting data increases, UIPE's improvement effects show a weakening trend. Specifically, in the Forget10, UIPE fails to improve the forgetting performance of KL Min., while it provides only slight improvements for the other three baseline unlearning methods. Baseline unlearning methods generally exhibit poor performance when handling large-scale target data \cite{maini2024tofu}, resulting in low-quality parameter update vectors $v$. Consequently, even though UIPE's amplifies $v$, it fails to significantly enhance the forgetting of related knowledge.

\subsection{Amplify Coefficient}
In UIPE, the amplify coefficient $\alpha$ controls additional parameter updates. We analyze the effect of different $\alpha$ on four unlearning methods using Forget01 dataset. For each method, we select an epoch as the base unlearning model and apply UIPE with varying $\alpha$ values. We then compare the forget quality of these UIPE models with that of the base model. When $\alpha = 0$, we measure the forget quality difference between the next epoch and base model.

\begin{figure}[th!]
\centering
\includegraphics[width=0.95\linewidth]{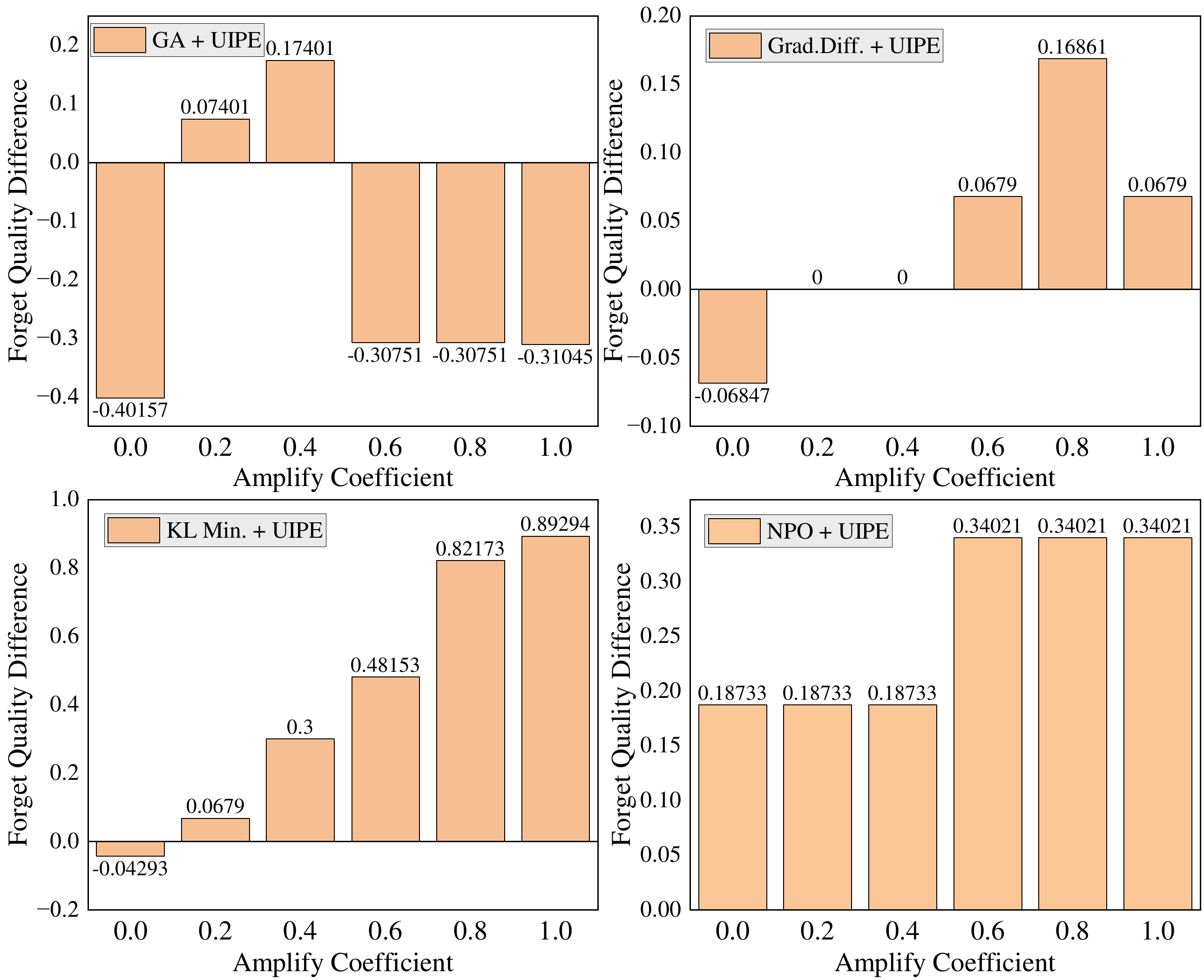}
\vspace{-5pt}
\caption{Performance of UIPE with different amplify coefficient $\alpha$.}
\label{amplify coefficient}
\end{figure}

As shown in Figure \ref{amplify coefficient}, in the Grad. Diff. method, larger $\alpha$ values improve forget quality. In the KL Min. method, forget quality consistently increases with rising $\alpha$ values. In the NPO method, forget quality exhibits relatively low sensitivity to changes in $\alpha$. For GA, forget quality first improves and then deteriorates as $\alpha$ increases, with the deterioration likely due to over-forgetting. As analyzed in Section \ref{UIPE method}, large $\alpha$ values may affect knowledge with low storage similarity, leading to a decline in model performance. However, the negative impact of UIPE on GA is still less severe than the decline observed in the original GA method.

\subsection{Forgetting Related knowledge}
Does UIPE effectively enhance the forgetting of related knowledge? As shown in Figure \ref{fig:k1,k2 test}, after the $8$th epoch, GA fails to further improve the forget quality of $\mathcal {P}_{\theta_{1}}$. Therefore, we choose to perform UIPE operations based on this.

\begin{figure}[th!]
\centering
\includegraphics[width=0.85\linewidth]{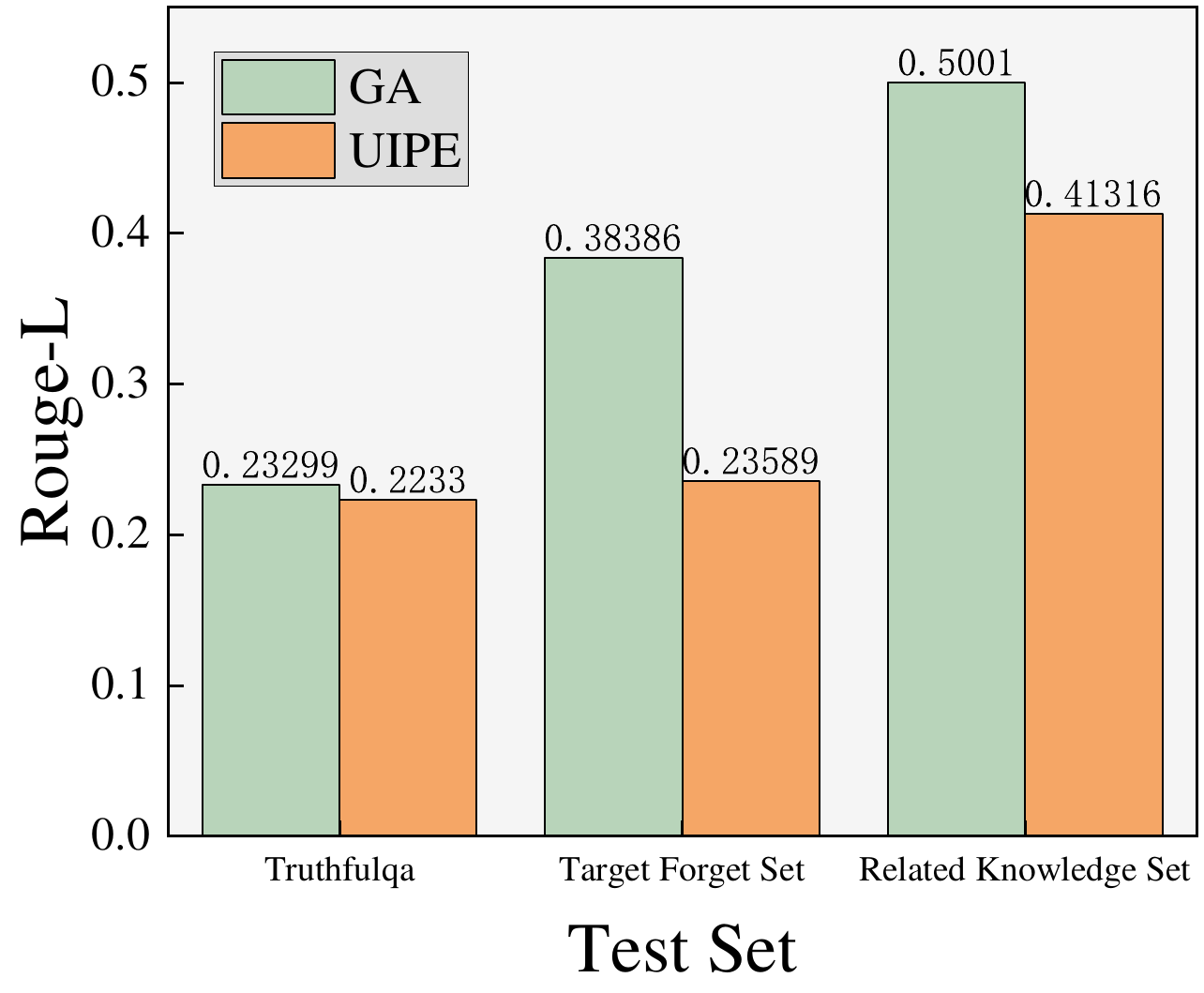}
\vspace{-5pt}
\caption{Performance changes after applying UIPE to the GA-trained model $\mathcal {P}_{\theta_{1}}$. A higher ROUGE-L score on TruthfulQA indicates better model utility, while lower ROUGE-L scores on the target forget set and related knowledge set indicate better forget quality.}
\label{forget related knowledge}
\end{figure}

As illustrated in Figure \ref{forget related knowledge}, while UIPE slightly reduces model utility, it significantly improves forget quality on both the related knowledge set and the target forget set. These results confirm that UIPE effectively facilitates the unlearning of related knowledge and strengthens the overall forgetting performance.

\section{Conclusion}
In this paper, we investigate the impact of knowledge related to forgetting targets on the effectiveness of target knowledge elimination. Based on this, we propose UIPE (Unlearning Improvement via Parameter Extrapolation), a technique that effectively forgets related knowledge without requiring additional training. Through extensive experimental validation across various unlearning methods, results demonstrate that UIPE significantly enhances these methods' ability to forget target knowledge.

\section*{Limitations}
Despite the effectiveness of our approach, there are two main limitations to be addressed in future work. First, The optimal amplify coefficient $\alpha$ requires manual selection across different baseline methods, necessitating further research to establish automated selection strategies for $\alpha$. Second, our study focuses on LLaMA2-7B. The larger parameter scales model (e.g., 70B)  typically contain richer and more complex knowledge representations. Further research is required to assess the effectiveness of UIPE on such larger-scale models.

\section*{Ethics Statement}
Our work aims to mitigate privacy and security concerns inherent in LLMs. However, users should exercise caution in practical applications, as alternative pathways may exist to expose unlearned knowledge. The existing datasets used in this study are obtained from official sources and utilized in accordance with their intended purposes. For newly created data, we strictly adhere to virtualization requirements during generation and employ manual verification to ensure no real information is disclosed, aligning with their intended use for public research and access.


\bibliography{reference}

\begin{thebibliography}{54}
\expandafter\ifx\csname natexlab\endcsname\relax\def\natexlab#1{#1}\fi

\bibitem[{Bourtoule et~al.(2021)Bourtoule, Chandrasekaran, Choquette-Choo, Jia, Travers, Zhang, Lie, and Papernot}]{bourtoule2021machine}
Lucas Bourtoule, Varun Chandrasekaran, Christopher~A Choquette-Choo, Hengrui Jia, Adelin Travers, Baiwu Zhang, David Lie, and Nicolas Papernot. 2021.
\newblock Machine unlearning.
\newblock In \emph{2021 IEEE Symposium on Security and Privacy (SP)}, pages 141--159. IEEE.

\bibitem[{Cao and Yang(2015)}]{cao2015towards}
Yinzhi Cao and Junfeng Yang. 2015.
\newblock Towards making systems forget with machine unlearning.
\newblock In \emph{2015 IEEE symposium on security and privacy}, pages 463--480. IEEE.

\bibitem[{Carlini et~al.(2021)Carlini, Tramer, Wallace, Jagielski, Herbert-Voss, Lee, Roberts, Brown, Song, Erlingsson et~al.}]{carlini2021extracting}
Nicholas Carlini, Florian Tramer, Eric Wallace, Matthew Jagielski, Ariel Herbert-Voss, Katherine Lee, Adam Roberts, Tom Brown, Dawn Song, Ulfar Erlingsson, et~al. 2021.
\newblock Extracting training data from large language models.
\newblock In \emph{30th USENIX Security Symposium (USENIX Security 21)}, pages 2633--2650.

\bibitem[{Chen and Yang(2023)}]{chen2023unlearn}
Jiaao Chen and Diyi Yang. 2023.
\newblock Unlearn what you want to forget: Efficient unlearning for llms.
\newblock \emph{arXiv preprint arXiv:2310.20150}.

\bibitem[{Chen et~al.(2023)Chen, Gao, Liu, Peng, and Wang}]{chen2023boundary}
Min Chen, Weizhuo Gao, Gaoyang Liu, Kai Peng, and Chen Wang. 2023.
\newblock Boundary unlearning: Rapid forgetting of deep networks via shifting the decision boundary.
\newblock In \emph{Proceedings of the IEEE/CVF Conference on Computer Vision and Pattern Recognition}, pages 7766--7775.

\bibitem[{Choi et~al.(2024)Choi, Park, Lee, and Choo}]{choi2024breaking}
Minseok Choi, ChaeHun Park, Dohyun Lee, and Jaegul Choo. 2024.
\newblock Breaking chains: Unraveling the links in multi-hop knowledge unlearning.
\newblock \emph{arXiv preprint arXiv:2410.13274}.

\bibitem[{Eldan and Russinovich(2023)}]{eldan2023s}
Ronen Eldan and Mark Russinovich. 2023.
\newblock Who's harry potter? approximate unlearning in llms.
\newblock \emph{arXiv preprint arXiv:2310.02238}.

\bibitem[{Fan et~al.(2024{\natexlab{a}})Fan, Liu, Hero, and Liu}]{fan2024challenging}
Chongyu Fan, Jiancheng Liu, Alfred Hero, and Sijia Liu. 2024{\natexlab{a}}.
\newblock Challenging forgets: Unveiling the worst-case forget sets in machine unlearning.
\newblock \emph{arXiv preprint arXiv:2403.07362}.

\bibitem[{Fan et~al.(2024{\natexlab{b}})Fan, Liu, Zhang, Wei, Wong, and Liu}]{fan2023salun}
Chongyu Fan, Jiancheng Liu, Yihua Zhang, Dennis Wei, Eric Wong, and Sijia Liu. 2024{\natexlab{b}}.
\newblock Salun: Empowering machine unlearning via gradient-based weight saliency in both image classification and generation.
\newblock In \emph{International Conference on Learning Representations}.

\bibitem[{Gandikota et~al.(2023)Gandikota, Materzynska, Fiotto-Kaufman, and Bau}]{gandikota2023erasing}
Rohit Gandikota, Joanna Materzynska, Jaden Fiotto-Kaufman, and David Bau. 2023.
\newblock Erasing concepts from diffusion models.
\newblock \emph{arXiv preprint arXiv:2303.07345}.

\bibitem[{Ginart et~al.(2019)Ginart, Guan, Valiant, and Zou}]{ginart2019making}
Antonio Ginart, Melody Guan, Gregory Valiant, and James~Y Zou. 2019.
\newblock Making ai forget you: Data deletion in machine learning.
\newblock \emph{Advances in neural information processing systems}, 32.

\bibitem[{Golatkar et~al.(2020)Golatkar, Achille, and Soatto}]{golatkar2020eternal}
Aditya Golatkar, Alessandro Achille, and Stefano Soatto. 2020.
\newblock Eternal sunshine of the spotless net: Selective forgetting in deep networks.
\newblock In \emph{Proceedings of the IEEE/CVF Conference on Computer Vision and Pattern Recognition}, pages 9304--9312.

\bibitem[{Graves et~al.(2021)Graves, Nagisetty, and Ganesh}]{graves2021amnesiac}
Laura Graves, Vineel Nagisetty, and Vijay Ganesh. 2021.
\newblock Amnesiac machine learning.
\newblock In \emph{Proceedings of the AAAI Conference on Artificial Intelligence}, volume~35, pages 11516--11524.

\bibitem[{Guo et~al.(2019)Guo, Goldstein, Hannun, and Van Der~Maaten}]{guo2019certified}
Chuan Guo, Tom Goldstein, Awni Hannun, and Laurens Van Der~Maaten. 2019.
\newblock Certified data removal from machine learning models.
\newblock \emph{arXiv preprint arXiv:1911.03030}.

\bibitem[{Hoofnagle et~al.(2019)Hoofnagle, Van Der~Sloot, and Borgesius}]{hoofnagle2019european}
Chris~Jay Hoofnagle, Bart Van Der~Sloot, and Frederik~Zuiderveen Borgesius. 2019.
\newblock The european union general data protection regulation: what it is and what it means.
\newblock \emph{Information \& Communications Technology Law}, 28(1):65--98.

\bibitem[{Huang et~al.(2022)Huang, Shao, and Chang}]{huang2022large}
Jie Huang, Hanyin Shao, and Kevin Chen-Chuan Chang. 2022.
\newblock Are large pre-trained language models leaking your personal information?
\newblock \emph{arXiv preprint arXiv:2205.12628}.

\bibitem[{Jang et~al.(2022)Jang, Yoon, Yang, Cha, Lee, Logeswaran, and Seo}]{jang2022knowledge}
Joel Jang, Dongkeun Yoon, Sohee Yang, Sungmin Cha, Moontae Lee, Lajanugen Logeswaran, and Minjoon Seo. 2022.
\newblock Knowledge unlearning for mitigating privacy risks in language models.
\newblock \emph{arXiv preprint arXiv:2210.01504}.

\bibitem[{Ji et~al.(2024)Ji, Liu, Zhang, Liu, Kompella, Liu, and Chang}]{ji2024reversing}
Jiabao Ji, Yujian Liu, Yang Zhang, Gaowen Liu, Ramana~Rao Kompella, Sijia Liu, and Shiyu Chang. 2024.
\newblock Reversing the forget-retain objectives: An efficient llm unlearning framework from logit difference.
\newblock \emph{arXiv preprint arXiv:2406.08607}.

\bibitem[{Jia et~al.(2023)Jia, Liu, Ram, Yao, Liu, Liu, Sharma, and Liu}]{jia2023model}
Jinghan Jia, Jiancheng Liu, Parikshit Ram, Yuguang Yao, Gaowen Liu, Yang Liu, Pranay Sharma, and Sijia Liu. 2023.
\newblock Model sparsity can simplify machine unlearning.
\newblock In \emph{Thirty-seventh Conference on Neural Information Processing Systems}.

\bibitem[{Jia et~al.(2024{\natexlab{a}})Jia, Liu, Zhang, Ram, Baracaldo, and Liu}]{jia2024wagle}
Jinghan Jia, Jiancheng Liu, Yihua Zhang, Parikshit Ram, Nathalie Baracaldo, and Sijia Liu. 2024{\natexlab{a}}.
\newblock Wagle: Strategic weight attribution for effective and modular unlearning in large language models.
\newblock \emph{arXiv preprint arXiv:2410.17509}.

\bibitem[{Jia et~al.(2024{\natexlab{b}})Jia, Zhang, Zhang, Liu, Runwal, Diffenderfer, Kailkhura, and Liu}]{jia2024soul}
Jinghan Jia, Yihua Zhang, Yimeng Zhang, Jiancheng Liu, Bharat Runwal, James Diffenderfer, Bhavya Kailkhura, and Sijia Liu. 2024{\natexlab{b}}.
\newblock Soul: Unlocking the power of second-order optimization for llm unlearning.
\newblock \emph{arXiv preprint arXiv:2404.18239}.

\bibitem[{Jin et~al.(2024)Jin, Cao, Wang, He, Yuan, Li, Chen, Liu, and Zhao}]{jin2024rwku}
Zhuoran Jin, Pengfei Cao, Chenhao Wang, Zhitao He, Hongbang Yuan, Jiachun Li, Yubo Chen, Kang Liu, and Jun Zhao. 2024.
\newblock Rwku: Benchmarking real-world knowledge unlearning for large language models.
\newblock \emph{arXiv preprint arXiv:2406.10890}.

\bibitem[{Kaplan et~al.(2020)Kaplan, McCandlish, Henighan, Brown, Chess, Child, Gray, Radford, Wu, and Amodei}]{kaplan2020scaling}
Jared Kaplan, Sam McCandlish, Tom Henighan, Tom~B Brown, Benjamin Chess, Rewon Child, Scott Gray, Alec Radford, Jeffrey Wu, and Dario Amodei. 2020.
\newblock Scaling laws for neural language models.
\newblock \emph{arXiv preprint arXiv:2001.08361}.

\bibitem[{Kumari et~al.(2023)Kumari, Zhang, Wang, Shechtman, Zhang, and Zhu}]{kumari2023ablating}
Nupur Kumari, Bingliang Zhang, Sheng-Yu Wang, Eli Shechtman, Richard Zhang, and Jun-Yan Zhu. 2023.
\newblock Ablating concepts in text-to-image diffusion models.
\newblock In \emph{Proceedings of the IEEE/CVF International Conference on Computer Vision}, pages 22691--22702.

\bibitem[{Kurmanji et~al.(2024)Kurmanji, Triantafillou, Hayes, and Triantafillou}]{kurmanji2024towards}
Meghdad Kurmanji, Peter Triantafillou, Jamie Hayes, and Eleni Triantafillou. 2024.
\newblock Towards unbounded machine unlearning.
\newblock \emph{Advances in neural information processing systems}, 36.

\bibitem[{Li et~al.(2024{\natexlab{a}})Li, Hsu, Marculescu et~al.}]{li2024machine}
Guihong Li, Hsiang Hsu, Radu Marculescu, et~al. 2024{\natexlab{a}}.
\newblock Machine unlearning for image-to-image generative models.
\newblock \emph{arXiv preprint arXiv:2402.00351}.

\bibitem[{Li et~al.(2024{\natexlab{b}})Li, Pan, Gopal, Yue, Berrios, Gatti, Li, Dombrowski, Goel, Phan et~al.}]{li2024wmdp}
Nathaniel Li, Alexander Pan, Anjali Gopal, Summer Yue, Daniel Berrios, Alice Gatti, Justin~D Li, Ann-Kathrin Dombrowski, Shashwat Goel, Long Phan, et~al. 2024{\natexlab{b}}.
\newblock The wmdp benchmark: Measuring and reducing malicious use with unlearning.
\newblock \emph{arXiv preprint arXiv:2403.03218}.

\bibitem[{Lin(2004)}]{lin-2004-rouge}
Chin-Yew Lin. 2004.
\newblock \href {https://aclanthology.org/W04-1013/} {{ROUGE}: A package for automatic evaluation of summaries}.
\newblock In \emph{Text Summarization Branches Out}, pages 74--81, Barcelona, Spain. Association for Computational Linguistics.

\bibitem[{Lin et~al.(2022)Lin, Hilton, and Evans}]{lin-etal-2022-truthfulqa}
Stephanie Lin, Jacob Hilton, and Owain Evans. 2022.
\newblock \href {https://doi.org/10.18653/v1/2022.acl-long.229} {{T}ruthful{QA}: Measuring how models mimic human falsehoods}.
\newblock In \emph{Proceedings of the 60th Annual Meeting of the Association for Computational Linguistics (Volume 1: Long Papers)}, pages 3214--3252, Dublin, Ireland. Association for Computational Linguistics.

\bibitem[{Liu et~al.(2024{\natexlab{a}})Liu, Wang, Flanigan, and Liu}]{liu2024large}
Chris~Yuhao Liu, Yaxuan Wang, Jeffrey Flanigan, and Yang Liu. 2024{\natexlab{a}}.
\newblock Large language model unlearning via embedding-corrupted prompts.
\newblock \emph{arXiv preprint arXiv:2406.07933}.

\bibitem[{Liu et~al.(2024{\natexlab{b}})Liu, Yao, Jia, Casper, Baracaldo, Hase, Xu, Yao, Li, Varshney et~al.}]{liu2024rethinking}
Sijia Liu, Yuanshun Yao, Jinghan Jia, Stephen Casper, Nathalie Baracaldo, Peter Hase, Xiaojun Xu, Yuguang Yao, Hang Li, Kush~R Varshney, et~al. 2024{\natexlab{b}}.
\newblock Rethinking machine unlearning for large language models.
\newblock \emph{arXiv preprint arXiv:2402.08787}.

\bibitem[{Liu et~al.(2024{\natexlab{c}})Liu, Zhang, Jaakkola, and Chang}]{liu2024revisiting}
Yujian Liu, Yang Zhang, Tommi Jaakkola, and Shiyu Chang. 2024{\natexlab{c}}.
\newblock Revisiting who’s harry potter: Towards targeted unlearning from a causal intervention perspective.
\newblock In \emph{Proceedings of the 2024 Conference on Empirical Methods in Natural Language Processing}, pages 8708--8731.

\bibitem[{Liu et~al.(2024{\natexlab{d}})Liu, Dou, Tan, Tian, and Jiang}]{liu2024towards}
Zheyuan Liu, Guangyao Dou, Zhaoxuan Tan, Yijun Tian, and Meng Jiang. 2024{\natexlab{d}}.
\newblock Towards safer large language models through machine unlearning.
\newblock \emph{arXiv preprint arXiv:2402.10058}.

\bibitem[{Liu et~al.(2023)Liu, Jiang, Shen, Peng, Lam, and Yuan}]{liu2023survey}
Ziyao Liu, Yu~Jiang, Jiyuan Shen, Minyi Peng, Kwok-Yan Lam, and Xingliang Yuan. 2023.
\newblock A survey on federated unlearning: Challenges, methods, and future directions.
\newblock \emph{arXiv preprint arXiv:2310.20448}.

\bibitem[{Maini et~al.(2024)Maini, Feng, Schwarzschild, Lipton, and Kolter}]{maini2024tofu}
Pratyush Maini, Zhili Feng, Avi Schwarzschild, Zachary~C Lipton, and J~Zico Kolter. 2024.
\newblock Tofu: A task of fictitious unlearning for llms.
\newblock \emph{arXiv preprint arXiv:2401.06121}.

\bibitem[{Nguyen et~al.(2022)Nguyen, Huynh, Ren, Nguyen, Liew, Yin, and Nguyen}]{nguyen2022survey}
Thanh~Tam Nguyen, Thanh~Trung Huynh, Zhao Ren, Phi~Le Nguyen, Alan Wee-Chung Liew, Hongzhi Yin, and Quoc Viet~Hung Nguyen. 2022.
\newblock A survey of machine unlearning.
\newblock \emph{arXiv preprint arXiv:2209.02299}.

\bibitem[{Qin et~al.(2024)Qin, Zhang, Han, Yu, Li, and Ji}]{qin2024does}
Jiaxin Qin, Zixuan Zhang, Chi Han, Pengfei Yu, Manling Li, and Heng Ji. 2024.
\newblock Why does new knowledge create messy ripple effects in llms?
\newblock In \emph{Proceedings of the 2024 Conference on Empirical Methods in Natural Language Processing}, pages 12602--12609.

\bibitem[{Rafailov et~al.(2024)Rafailov, Sharma, Mitchell, Manning, Ermon, and Finn}]{rafailov2024direct}
Rafael Rafailov, Archit Sharma, Eric Mitchell, Christopher~D Manning, Stefano Ermon, and Chelsea Finn. 2024.
\newblock Direct preference optimization: Your language model is secretly a reward model.
\newblock \emph{Advances in Neural Information Processing Systems}, 36.

\bibitem[{Rosen(2011)}]{rosen2011right}
Jeffrey Rosen. 2011.
\newblock The right to be forgotten.
\newblock \emph{Stan. L. Rev. Online}, 64:88.

\bibitem[{Sekhari et~al.(2021)Sekhari, Acharya, Kamath, and Suresh}]{sekhari2021remember}
Ayush Sekhari, Jayadev Acharya, Gautam Kamath, and Ananda~Theertha Suresh. 2021.
\newblock Remember what you want to forget: Algorithms for machine unlearning.
\newblock \emph{Advances in Neural Information Processing Systems}, 34:18075--18086.

\bibitem[{Shumailov et~al.(2024)Shumailov, Hayes, Triantafillou, Ortiz-Jimenez, Papernot, Jagielski, Yona, Howard, and Bagdasaryan}]{shumailov2024ununlearning}
Ilia Shumailov, Jamie Hayes, Eleni Triantafillou, Guillermo Ortiz-Jimenez, Nicolas Papernot, Matthew Jagielski, Itay Yona, Heidi Howard, and Eugene Bagdasaryan. 2024.
\newblock Ununlearning: Unlearning is not sufficient for content regulation in advanced generative ai.
\newblock \emph{arXiv preprint arXiv:2407.00106}.

\bibitem[{Touvron et~al.(2023)Touvron, Martin, Stone, Albert, Almahairi, Babaei, Bashlykov, Batra, Bhargava, Bhosale et~al.}]{touvron2023llama}
Hugo Touvron, Louis Martin, Kevin Stone, Peter Albert, Amjad Almahairi, Yasmine Babaei, Nikolay Bashlykov, Soumya Batra, Prajjwal Bhargava, Shruti Bhosale, et~al. 2023.
\newblock Llama 2: Open foundation and fine-tuned chat models.
\newblock \emph{arXiv preprint arXiv:2307.09288}.

\bibitem[{Ullah et~al.(2021)Ullah, Mai, Rao, Rossi, and Arora}]{ullah2021machine}
Enayat Ullah, Tung Mai, Anup Rao, Ryan~A Rossi, and Raman Arora. 2021.
\newblock Machine unlearning via algorithmic stability.
\newblock In \emph{Conference on Learning Theory}, pages 4126--4142. PMLR.

\bibitem[{Wang et~al.(2022)Wang, Guo, Xie, and Qi}]{wang2022federated}
Junxiao Wang, Song Guo, Xin Xie, and Heng Qi. 2022.
\newblock Federated unlearning via class-discriminative pruning.
\newblock In \emph{Proceedings of the ACM Web Conference 2022}, pages 622--632.

\bibitem[{Wang et~al.(2023)Wang, Chen, Yuan, Zeng, Wong, and Yin}]{wang2023kga}
Lingzhi Wang, Tong Chen, Wei Yuan, Xingshan Zeng, Kam-Fai Wong, and Hongzhi Yin. 2023.
\newblock Kga: A general machine unlearning framework based on knowledge gap alignment.
\newblock \emph{arXiv preprint arXiv:2305.06535}.

\bibitem[{Wei et~al.(2022)Wei, Tay, Bommasani, Raffel, Zoph, Borgeaud, Yogatama, Bosma, Zhou, Metzler et~al.}]{wei2022emergent}
Jason Wei, Yi~Tay, Rishi Bommasani, Colin Raffel, Barret Zoph, Sebastian Borgeaud, Dani Yogatama, Maarten Bosma, Denny Zhou, Donald Metzler, et~al. 2022.
\newblock Emergent abilities of large language models.
\newblock \emph{arXiv preprint arXiv:2206.07682}.

\bibitem[{Xie et~al.(2024)Xie, Fang, Pi, and Gong}]{xie2024gradsafe}
Yueqi Xie, Minghong Fang, Renjie Pi, and Neil Gong. 2024.
\newblock Gradsafe: Detecting jailbreak prompts for llms via safety-critical gradient analysis.
\newblock In \emph{Proceedings of the 62nd Annual Meeting of the Association for Computational Linguistics (Volume 1: Long Papers)}, pages 507--518.

\bibitem[{Yao et~al.(2024)Yao, Chien, Du, Niu, Wang, Cheng, and Yue}]{yao2024machine}
Jin Yao, Eli Chien, Minxin Du, Xinyao Niu, Tianhao Wang, Zezhou Cheng, and Xiang Yue. 2024.
\newblock Machine unlearning of pre-trained large language models.
\newblock \emph{arXiv preprint arXiv:2402.15159}.

\bibitem[{Yao et~al.(2023)Yao, Xu, and Liu}]{yao2023large}
Yuanshun Yao, Xiaojun Xu, and Yang Liu. 2023.
\newblock Large language model unlearning.
\newblock \emph{arXiv preprint arXiv:2310.10683}.

\bibitem[{Yu et~al.(2023)Yu, Jeoung, Kasi, Yu, and Ji}]{yu2023unlearning}
Charles Yu, Sullam Jeoung, Anish Kasi, Pengfei Yu, and Heng Ji. 2023.
\newblock Unlearning bias in language models by partitioning gradients.
\newblock In \emph{Findings of the Association for Computational Linguistics: ACL 2023}, pages 6032--6048.

\bibitem[{Zhang et~al.(2023{\natexlab{a}})Zhang, Finckenberg-Broman, Hoang, Pan, Xing, Staples, and Xu}]{zhang2023right}
Dawen Zhang, Pamela Finckenberg-Broman, Thong Hoang, Shidong Pan, Zhenchang Xing, Mark Staples, and Xiwei Xu. 2023{\natexlab{a}}.
\newblock Right to be forgotten in the era of large language models: Implications, challenges, and solutions.
\newblock \emph{arXiv preprint arXiv:2307.03941}.

\bibitem[{Zhang et~al.(2023{\natexlab{b}})Zhang, Wang, Xu, Wang, and Shi}]{zhang2023forget}
Eric Zhang, Kai Wang, Xingqian Xu, Zhangyang Wang, and Humphrey Shi. 2023{\natexlab{b}}.
\newblock Forget-me-not: Learning to forget in text-to-image diffusion models.
\newblock \emph{arXiv preprint arXiv:2303.17591}.

\bibitem[{Zhang et~al.(2024)Zhang, Lin, Bai, and Mei}]{zhang2024negative}
Ruiqi Zhang, Licong Lin, Yu~Bai, and Song Mei. 2024.
\newblock Negative preference optimization: From catastrophic collapse to effective unlearning.
\newblock \emph{arXiv preprint arXiv:2404.05868}.

\bibitem[{Zhao et~al.(2024)Zhao, Ding, Jia, Wang, and Qian}]{zhao2024gender}
Jinman Zhao, Yitian Ding, Chen Jia, Yining Wang, and Zifan Qian. 2024.
\newblock Gender bias in large language models across multiple languages.
\newblock \emph{arXiv preprint arXiv:2403.00277}.

\end{thebibliography}
\bibliographystyle{acl_natbib}

\onecolumn
\clearpage
\appendix

\section{Prompt and Data Sample} \label{Prompt and Data Sample}
Table \ref{tab:prompt} illustrates the data construction prompt used in our preliminary experiments, which requests GPT-4o to generate information for 12 virtual individuals. The information for each virtual individual consists of 10 specific attributes, with each attribute containing two question-answer pairs: K1 and K2. Based on the K2 question-answer pairs and the general common-sense knowledge of the large model, it is possible to infer the K1 question-answer pairs, indicating a logical relationship between them. Table \ref{tab:data_sample} presents a specific example of one generated virtual individual. After generating the 12 virtual individuals, we compile all K1 question-answer pairs into the target forget set, while all K2 question-answer pairs form the related knowledge set.

\begin{table*}[h]
    \centering
    \caption{The prompt for generating fictional individuals data using GPT-4o.}
\resizebox{\linewidth}{!}
{\begin{tcolorbox}[
    colback=gray!5,
    colframe=gray!40,
    width=\textwidth,
    title=Prompt - One Shot
]
Please provide 12 sample in JSON format, each sample
containing information about a person, including 10 Q\&A
pairs K1 and K2, with the following requirements:\\

1. Each question and answer pair is about personal
privacy (biometric features, occupational information,
address, health and medical information, personal
property information, track of when, network activity
information, family, marriage, personal communication
information), and all information presented here is
fictional.\\

2. Based on K2, K1 can be inferred through common
sense.\\

3. Strictly follow the privacy type of each question and
answer in the example, but the questions should have
diversity.\\

Here is an example about John:\\

......

\end{tcolorbox}}  
    \label{tab:prompt}
\end{table*}

\begin{table*}[t!]
    \centering
    \caption{Sample of fictional individual generated based on 10 specific attributes.}
\resizebox{\linewidth}{!}
{\begin{tcolorbox}[
    colback=gray!5,
    colframe=gray!40,
    width=\textwidth,
    title=Sample Data
]
\begin{lstlisting}
{
    "name": "john", 
    "Biometric features": [ 
        {
            "type": "k1", 
            "question": "What is John's blood type?", 
            "answer": "John has type A positive blood." 
        }, 
        {
            "type": "k2", 
            "question": "What blood types can John donate to?", 
            "answer": "John can donate blood to type A, AB, and O positive individuals." 
       }
    ], 
    "Occupational information": [
        {
            "type": "k1",
            "question": "What is John's profession?",
            "answer": "John works as a software engineer at a tech company."
        },
        {
            "type": "k2",
            "question": "What programming languages does John use at work?",
            "answer": "John primarily uses Python, Java, and JavaScript in his daily work."
        }
    ], 
    "Address": [
        {
            "type": "k1",
            "question": "Where does John live?",
            "answer": "John lives in a townhouse in a suburban neighborhood."
        },
        {
            "type": "k2",
            "question": "How is John's living environment?",
            "answer": "John's home has good air quality away from the bustle of downtown, with a small yard and terrace."
        }
    ], 
    "Health and medical information": [
        {
            "type": "k1",
            "question": "Does John have any chronic conditions?",
            "answer": "John has been diagnosed with asthma."
        },
        {
            "type": "k2",
            "question": "What medication does John use?",
            "answer": "John uses an inhaler with a steroid medication."
        }
    ]
    ...
}
\end{lstlisting}
\end{tcolorbox}}  
    \label{tab:data_sample}
\end{table*}

\section{Training Details}
We select LLaMA-2-7b-chat as our base model and employ Low-Rank Adaptation (LoRA) for both fine-tuning and unlearning processes. During the fine-tuning phase, we set the learning rate to 1e-4, batch size to 4, and LoRA rank to 4, conducting training on a single NVIDIA RTX 4090 GPU. For the unlearning phase, we adjust the learning rate to 5e-5 while maintaining the batch size of 4 and LoRA rank of 4, also training on a single NVIDIA RTX 4090 GPU. In both phases, we exclusively update the parameters of two target modules: "q\_proj" and "v\_proj".

\section{Algorithm}
\begin{algorithm}[H]
\caption{UIPE}
\label{alg:uipe}
\begin{algorithmic}[1]
\Require
\Statex Initial model parameters $\theta_{\text{ini}}$
\Statex Target forget dataset $\mathcal{D}_f$
\Statex Training epochs $T$
\Statex Extrapolation coefficient $\alpha$
\Ensure
\Statex Enhanced unlearned model $\theta_{\text{uipe}}$

\Procedure{Unlearning Phase}{}
    \For{$t = 1$ \textbf{to} $T$}
        \State $\theta_t \gets \theta_{t-1} + \eta\nabla_\theta[\mathcal{L}_{GA}(\theta)]$ \Comment{Initial forgetting training}
        \State $U_t \gets \text{EvalUtility}(\theta_t, \mathcal{D}_r)$
        \State $F_t \gets \text{EvalQuality}(\theta_t, \mathcal{D}_f)$
    \EndFor
    \State $\theta_{\text{un}} \gets select_{\theta_t} [F_t,U_t]$ \Comment{Select a model that balances forget quality and model utility}
\EndProcedure

\State \textbf{Update Vector Calculation:}
\State $v \gets \theta_{\text{un}} - \theta_{\text{ini}}$ \Comment{Calculate update vector}

\State \textbf{Knowledge Extrapolation:}
\State $\theta_{\text{uipe}} \gets \theta_{\text{un}} + \alpha \cdot v$ \Comment{Parameter extrapolation}

\State \Return $\theta_{\text{uipe}}$
\end{algorithmic}
\end{algorithm}

\section{Experimental details}
\subsection{Baseline LLM unlearning methods} \label{Baseline LLM unlearning methods}
In addition to the basic Gradient Ascent (GA) method, we also conduct experiments on three other unlearning techniques using the TOFU benchmark
\begin{itemize}
    \item \textbf{Grad. Diff.} This approach not only aims to increase the loss on the forget dataset $\mathcal{D}_{f}$ but also strives to maintain performance on the retain dataset $\mathcal{D}_{r}$.
\end{itemize}
\begin{itemize}
    \item \textbf{KL Min.} This approach not only seeks to increase the loss on the forget dataset $\mathcal{D}_{f}$ but also minimizes the Kullback-Leibler (KL) divergence between the fine-tune model and the unlearning model on the retain dataset $\mathcal{D}_{r}$.
\end{itemize}
\begin{itemize}
    \item \textbf{NPO} Inspired by preference optimization, this approach can be regarded as a variant that focuses solely on negative samples.
\end{itemize}
\subsection{Training Details}
In the TOFU benchmark, The authors provide the tofu\_ft\_llama2-7b model, which is fine-tuned on the TOFU dataset using LLaMA-2-7b-chat as the base model. We use this model for our experiments. We refer to the experimental details of TOFU and NPO for full fine-tuning. Specifically, we employ a learning rate of 1e-5 for the Forget01 and Forget05 datasets, and a learning rate of 1e-6 for the Forget10 dataset, aiming to maximize the performance of these baseline methods. During training, the batch size is set to 1, and the process is conducted on two NVIDIA A800 80GB GPUs.

\end{document}